%% file: main.tex
\documentclass[letterpaper]{article} 
\usepackage{aaai2027}  
\usepackage[hyphens]{url}  
\usepackage{graphicx} 
\usepackage{natbib}  
\usepackage{caption} 
\usepackage{algorithm}
\usepackage{algorithmic}

\usepackage{newfloat}
\usepackage{listings}
\DeclareCaptionStyle{ruled}{labelfont=normalfont,labelsep=colon,strut=off} 
\floatstyle{ruled}
\newfloat{listing}{tb}{lst}{}
\floatname{listing}{Listing}

\usepackage{booktabs}
\usepackage{amsmath}
\usepackage{amssymb}
\usepackage{multirow}
\usepackage{xcolor}
\usepackage{tcolorbox}
\tcbuselibrary{breakable}   
\tcbuselibrary{skins}       

\nocopyright 

\title{GraftSR: Grafting Authentic Textures for Real-World Image Super-Resolution \\via Identical-Instance Guidance}
\author{
    Qifan Yu\textsuperscript{\rm 1},
    Haoran Bai\textsuperscript{\rm 1},
    Zongyao He\textsuperscript{\rm 1},
    Weijie He\textsuperscript{\rm 1},
    Sibin Deng\textsuperscript{\rm 1},
    Honggang Qi\textsuperscript{\rm 2},
    Ying Chen\textsuperscript{\rm 1}\corresponding,
}
\affiliations{
    \textsuperscript{\rm 1}Taobao \& Tmall Group of Alibaba,
    \textsuperscript{\rm 2}University of Chinese Academy of Sciences

    \{yuqifan.yqf, baihaoran.bhr\}@taobao.com\\[0.6em]
    {\large\textcolor{magenta}{\url{https://yuqifan1117.github.io/GraftSR/}}}
}

\begin{document}

\maketitle

\begin{abstract}
Diffusion-based real-world image super-resolution (SR) achieves impressive perceptual quality but inherently suffers from severe texture hallucination.
To overcome this limitation, we propose \emph{GraftSR}, a texture-reference-guided generative SR framework that leverages reference images of the identical instance to anchor the restoration of authentic textures.
However, severe spatial misalignment between low-quality inputs and their references poses significant challenges, often leading to ambiguous transfer targets and background feature leakage.
To address these issues, GraftSR employs a novel \emph{dual-mask reference guidance} mechanism that systematically decouples the cross-view texture injection process.
By explicitly isolating \emph{what} authentic textures to extract from the reference and precisely localizing \emph{where} to apply them within the target, GraftSR achieves robust texture transfer without relying on brittle spatial alignment.
Furthermore, to bridge the critical gap in appropriate training data, we construct \emph{TexRefSR-141K}, the first large-scale dataset providing high-quality reference tuples equipped with complementary spatial masks.
Extensive experiments on our newly established benchmark, \emph{TexRefSR-Eval}, demonstrate that GraftSR sets a new state-of-the-art. Notably, it reduces LPIPS by 20.2\% over top-performing baselines, achieving superior reference-faithful restoration.

\end{abstract}

\section{Introduction}
\begin{figure*}[!t]
    \centering
    \includegraphics[width=0.9\linewidth]{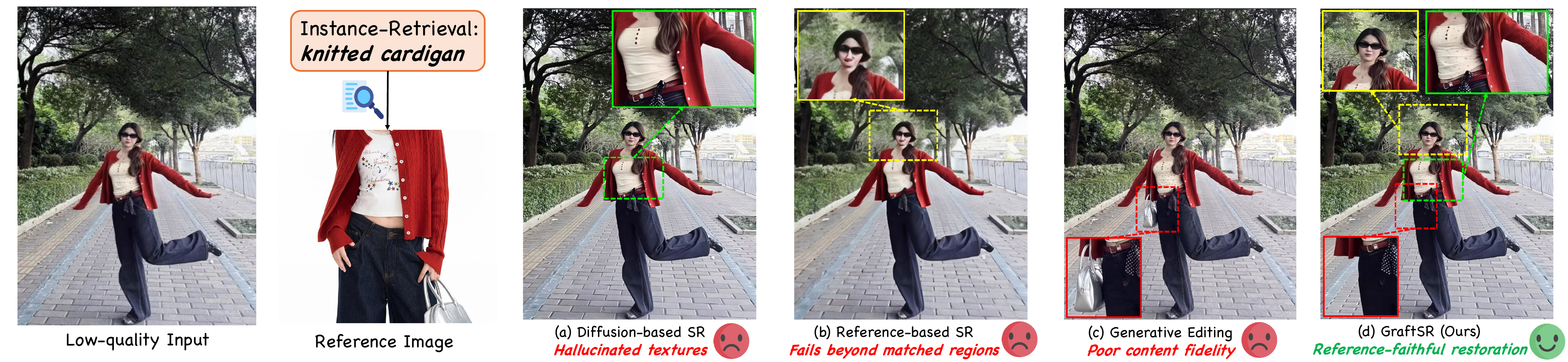}
    \caption{
    \textbf{GraftSR} achieves reference-faithful image SR of fine-grained details~(\textit{e.g.}, material textures and delicate prints), whereas existing methods exhibit distinct limitations: (a)~Diffusion-based SR generates \textit{hallucinated textures}; (b)~Reference-based SR introduces \textit{artifacts beyond matched regions}; and (c)~Generative editing model suffers from \textit{poor content fidelity}.
}
  \label{teaser}
\end{figure*}

Diffusion-based real-world image super-resolution (SR)~\cite{wu2024one, duan2025dit4sr, fang2026one} aims to restore high-quality (HQ) images from degraded observations by harnessing the immense potential of generative models~\cite{wu2025qwen, peebles2023scalable, labs2025flux, podell2024sdxl}, achieving perceptual quality far beyond that of earlier CNN-based approaches~\cite{zhang2017learning, soh2022variational, cui2024revitalizing}. However, a fundamental limitation persists: \textit{the generative restoration process inherently hallucinates textures that deviate from their authentic appearance} due to the ill-posed nature of SR~(see Fig.~\ref{teaser}(a)). This severe hallucination destroys the fine-grained identity of the subject, strictly limiting its deployment in real-world scenarios that demand precise physical fidelity (\textit{e.g.}, e-commerce applications).

This limitation motivates the integration of reference images to anchor the generative process, aiming to restore a low-quality (LQ) input into a HQ output whose textures are faithful to a reference of the identical instance.
Recently, generative editing approaches (\textit{e.g.}, Gemini-3-Pro) have emerged that can formally incorporate reference images to transfer customized textures into target regions.
However, while these methods can achieve texture transfer, they often do so at the expense of the original geometric structure, resulting in poor content fidelity~(see Fig.~\ref{teaser}(c)).
Therefore, it is of great interest to develop an effective texture-reference-guided generative SR framework that seamlessly transfers authentic reference textures while strictly preserving the structural and content consistency of the input image.

However, realizing this goal is highly challenging due to the severe cross-view spatial misalignments~(\textit{e.g.}, varying poses, scales, and viewpoints) inherent in real-world identical-instance images.
Existing reference-based SR methods rely heavily on brittle feature-level matching, suffering from severe artifacts beyond locally matched regions under such spatial misalignments~(see Fig.~\ref{teaser}(b)).
Thus, achieving faithful cross-view texture transfer faces two intertwined challenges:
\textit{(i)} background feature leakage, since irrelevant reference contexts make it hard to isolate exactly \emph{what to reference};
and \textit{(ii)} ambiguous transfer target, since it is difficult to precisely localize \emph{where to transfer} the reference textures.

To overcome these challenges, we present \emph{GraftSR}, a novel generative framework driven by a \emph{dual-mask reference guidance} mechanism.
Built upon MMDiT backbone~\cite{wu2025qwen}, it systematically tackles the aforementioned bottlenecks by explicitly injecting reference textures through two synergistic condition tokens:
\textit{(i)} mask-modulated reference tokens that isolate valid reference regions to explicitly determine \emph{what} authentic textures to extract;
and \textit{(ii)} region-aware semantic tokens that precisely localize \emph{where} the texture should be applied within the target.
By jointly attending to these tokens within a unified sequence, GraftSR implicitly learns robust cross-view texture aggregation without relying on brittle spatial matching.
Finally, an adversarial training objective~\cite{fang2026one} distills the generative process for efficient one-step image SR framework.

Furthermore, to address the critical absence of identical-instance training data, we curate \emph{TexRefSR-141K}, the first large-scale dataset tailored for this task.
Through an automated data construction pipeline, we pair images of strictly identical instances from multi-view galleries and supplement them with explicit complementary spatial masks.
Specifically, the reference mask isolates the authentic texture source, while the target mask demarcates the exact restoration region.
Comprising over 141K high-quality tuples across 61K distinct instances, this dataset provides the essential spatial selectivity to decouple textures from complex backgrounds, establishing a vital data foundation for the community.

Extensive experiments on our newly established \emph{TexRefSR-Eval} benchmark, which enables faithful evaluation against identical-instance references, validate the superiority of GraftSR against state-of-the-art baselines.
Quantitatively, a notable 20.2\% reduction in LPIPS over top-performing methods confirms that the generated details are faithfully anchored to reference textures rather than arbitrarily hallucinated.
Qualitatively, GraftSR achieves highly precise cross-view texture transfer while strictly preserving the original geometric structures, delivering reference-faithful restoration (see Fig.~\ref{teaser}(d)). In summary, our main contributions are:

\begin{itemize}
    \item We propose \emph{GraftSR}, a novel texture-reference-guided generative SR framework that suppresses generative hallucinations and achieves highly faithful texture transfer from cross-view identical-instance references.
    \item We construct \emph{TexRefSR-141K} and \emph{TexRefSR-Eval}, the first large-scale dataset and benchmark suite providing high-quality identical-instance tuples equipped with complementary spatial masks.
    \item Extensive experiments demonstrate that GraftSR establishes a new state-of-the-art in reference-guided SR, bridging the gap between perceptual quality and texture fidelity for faithful texture restoration.
\end{itemize}

\section{Related Work}

\subsection{Diffusion-based Image Restoration}
Recent advances in diffusion-based generative models~\cite{rombach2022high, esser2024scaling, wu2025qwen} have shifted image restoration from pixel-level regression~\cite{zhang2018image, liang2021swinir} to perceptual synthesis. Early multi-step approaches have successfully adapted large-scale diffusion models for image super-resolution~\cite{yu2024scaling, duan2025dit4sr}. To mitigate the heavy computational overhead, one-step distillation models have recently emerged. Specifically, OSEDiff~\cite{wu2024one} leverages variational score distillation, while PiSA-SR~\cite{sun2025pixel} and ODTSR~\cite{fang2026one} decouple generation branches to optimize perceptual quality and semantic control simultaneously. To capture finer visual details, VOSR~\cite{wu2026vosr} departs from standard T2I paradigms by proposing a vision-only framework that substitutes textual prompts with deep visual features extracted from pretrained encoders. Despite their impressive performance, these blind restoration methods lack explicit target texture priors, often resulting in spurious artifacts or hallucinated details. In contrast, our work leverages reference images of the identical instance as precise texture anchors to faithfully transfer reference textures during generative restoration.

\subsection{Reference-based Image Restoration}

To bypass the ill-posed issues of single-image paradigms, reference-based super-resolution~(RefSR) restores high-fidelity details based on auxiliary images. Early matching-based methods~\cite{zhang2019image, yang2020ttsr, zhang2022rrsr, jiang2021c2matching, cao2022reference, zhang2023lmr, lu2021masa} and advanced weighting strategies~\cite{guo2024refir, varanka2024pfstorer} heavily rely on localized patch alignment. Although recent diffusion-based RefSR models~\cite{wang2026trust, zhao2026garmentzoom} attempt to process unaligned references, deriving reference pairs from semantically approximate images or single-image crops inherently restricts them to global feature matching. Consequently, they inevitably suffer from background leakage when facing unaligned references. In contrast, our method achieves robust texture transfer via dual-mask reference guidance.

\section{Method}
\begin{figure*}[!t]
    \centering
    \includegraphics[width=0.9\linewidth]{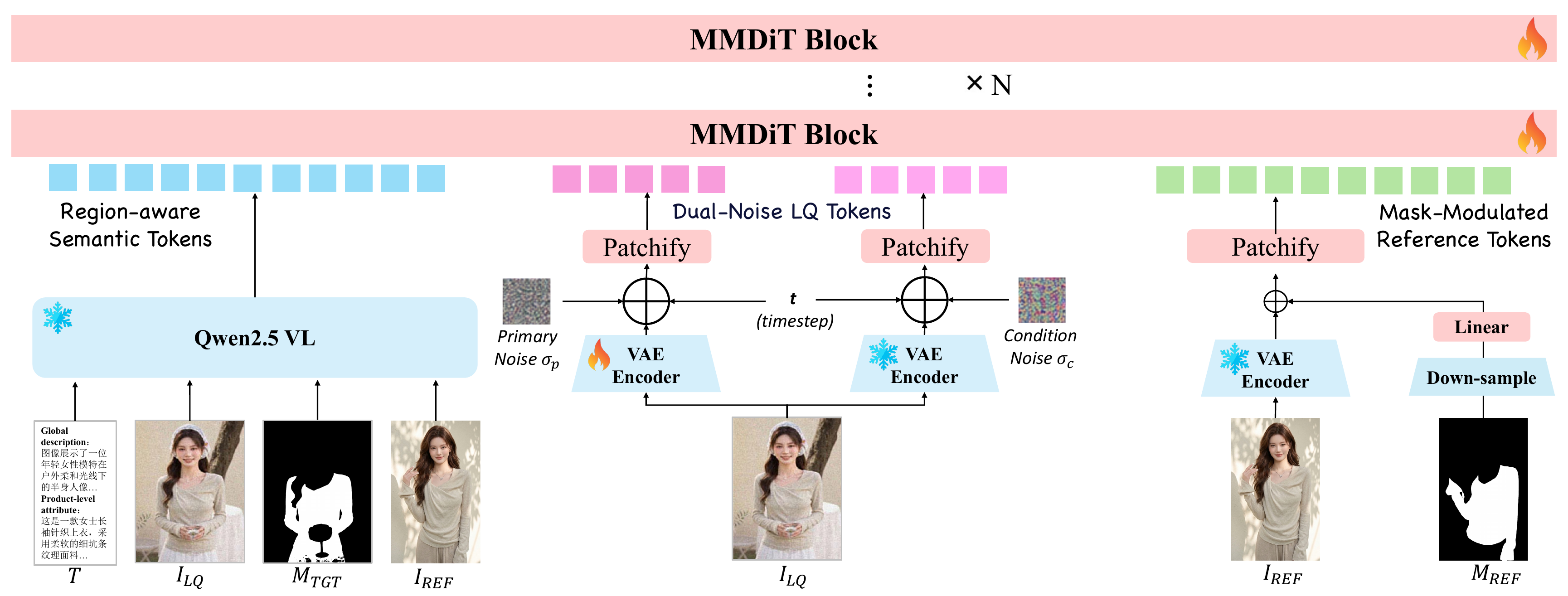}
    \caption{Overview architecture of GraftSR. Driven by a \emph{dual-mask reference guidance} mechanism, it extracts \emph{mask-modulated reference tokens} and \emph{region-aware semantic tokens} to condition the MMDiT blocks for one-step reference-faithful restoration.}
  \label{pipeline}
\end{figure*}

Unlike conventional blind super-resolution, our texture-reference-guided SR framework aims to reconstruct degraded observations by faithfully transferring authentic textures from identical-instance reference images.
Given a LQ input $I_{LQ}$ and a HQ texture reference $I_{REF}$ of the identical instance, the goal is to synthesize a HQ output $\hat{I}_{HQ}$ that strictly preserves the spatial structure of $I_{LQ}$ while achieving precise texture fidelity to $I_{REF}$.

\subsection{Dual-Mask Reference-Guided Architecture} 
To achieve authentic texture fidelity without relying on brittle spatial alignment, we propose \emph{GraftSR}, introducing a novel \emph{dual-mask reference guidance} mechanism to explicitly resolve the critical ambiguities of \emph{what to reference} and \emph{where to transfer}.
Built upon the MMDiT backbone~\cite{wu2025qwen}, we design two synergistic conditioning tokens, namely \emph{mask-modulated reference tokens} and \emph{region-aware semantic tokens}, to effectively overcome these two challenges.
As illustrated in Fig.~\ref{pipeline}, these extracted tokens jointly condition the cascaded MMDiT blocks to denoise the dual-noise latent, which is ultimately distilled via a one-step adversarial objective~\cite{fang2026one} for efficient image SR.

\paragraph{Semantic and Spatial Condition Extraction.} 
To explicitly establish the semantic and spatial correspondences of the identical instance between the LQ input $I_{LQ}$ and the HQ reference $I_{REF}$, we employ an automated extraction process prior to the diffusion model. First, we leverage a Vision-Language Model (VLM)~\cite{bai2025qwen3} to jointly analyze both images. The VLM identifies the semantic name of the shared instance and generates a comprehensive text caption $T$ for $I_{LQ}$, which acts as an optional textual condition for the MMDiT backbone. Subsequently, using the identified instance name as a prompting cue, we employ SAM3~\cite{carion2025sam} to perform zero-shot segmentation. This step extracts the target region mask $M_{TGT}$ from $I_{LQ}$ and the reference mask $M_{REF}$ from $I_{REF}$, explicitly localizing the identical instance across both views. These extracted masks and the caption seamlessly serve as the fundamental conditions for the subsequent token generation.

\paragraph{Dual-Mask Synergistic Tokens.} 
To explicitly decouple the cross-view texture injection process, we formulate two synergistic conditioning tokens driven by the extracted masks. First, to determine \textit{what to reference}, we encode $I_{REF}$ using a frozen VAE. Since unaligned identical-instance references inevitably contain irrelevant backgrounds, we modulate the reference latent with the corresponding spatial mask to explicitly isolate valid texture-bearing regions:
\begin{equation}
    z_{\text{ref}} = \mathcal{E}_{\text{VAE}}(I_{REF}) + \phi(\mathcal{D}(M_{REF})),
\end{equation}
where $z_{\text{ref}}$ denotes the mask-modulated reference tokens, $\mathcal{E}_{\text{VAE}}$ denotes the frozen VAE encoder, $\mathcal{D}$ is the down-sampling operation, and $\phi(\cdot)$ is a learnable projection. Rather than applying a hard mask that might disrupt the continuous latent space, this additive modulation softly intervenes in the feature space, isolating authentic textures while preserving the model's general generative capability.

Second, to accurately designate \textit{where to transfer} these extracted textures, we establish a high-level spatial-semantic understanding of the target. We extract multi-modal condition tokens via the frozen Qwen2.5-VL encoder~\cite{bai2025qwen3}:
\begin{equation}
    z_{\text{sem}} = \mathcal{E}_{\text{VLM}}(I_{LQ},\; M_{TGT},\; I_{REF},\; T),
\end{equation}
where $\mathcal{E}_{\text{VLM}}$ denotes the Qwen2.5-VL encoder. By jointly embedding the text prompt $T$ with these visual-spatial clues, $\mathcal{E}_{\text{VLM}}$ yields region-aware semantic tokens $z_{\text{sem}}$. Together, $z_{\text{ref}}$ and $z_{\text{sem}}$ form a complementary pair, precisely guiding the network to transfer the right textures to the right locations.

\paragraph{Model Training.} 
Following~\cite{fang2026one}, we adopt a dual-noise strategy to optimally balance perceptual quality and texture fidelity. Specifically, dual-level noises are injected into the latent representation of the LQ input to formulate the noisy target tokens.
To train the MMDiT backbone as an efficient one-step SR model, these noisy tokens are concatenated along the sequence dimension with our explicitly extracted conditions and subsequently processed by the MMDiT blocks.
The predicted denoised latent is then decoded back to the pixel space, and the model is optimized using a combination of MSE, DISTS~\cite{ding2020dists}, and an adversarial objectives.
For adversarial supervision, we employ a relativistic GAN loss driven by a Wan2.1-1.3B-based discriminator~\cite{wan2025wan} to guarantee high-fidelity perceptual generation.
More training details are deferred to the \textit{Supplementary Materials}.

\begin{figure}[!t]
    \centering
    \includegraphics[width=0.95\linewidth]{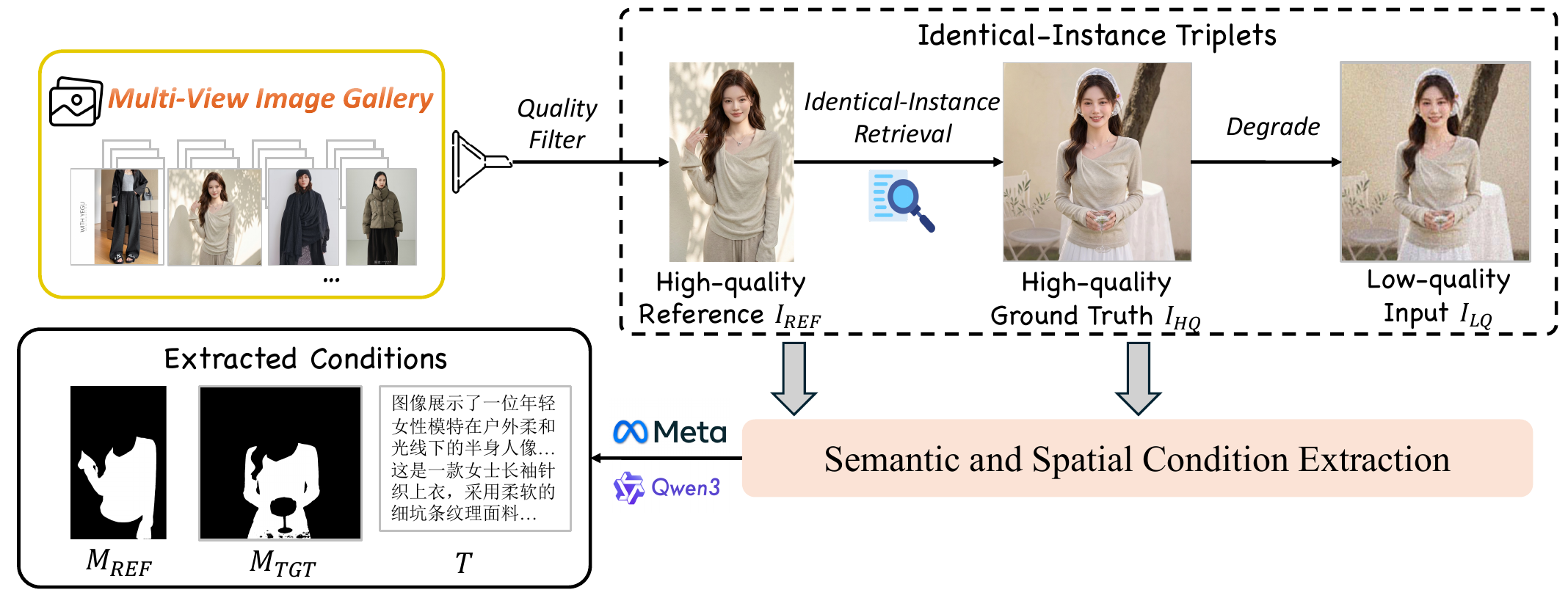}
    \caption{The proposed VLM-driven data curation pipeline. It retrieves identical-instance triplets from multi-view galleries and extracts the complementary masks and unified caption to support authentic texture transfer.} 
    \label{fig:data_pipeline}
\end{figure}

\subsection{Identical-Instance Dataset Construction}
To effectively train GraftSR, the availability of large-scale, strictly identical-instance data is imperative.
Prevailing reference-based SR datasets merely pair semantically approximate images, inevitably synthesizing textures that deviate from the target instance's authentic appearance.
To fundamentally shift this paradigm, we exploit large-scale e-commerce multi-view image galleries, which uniquely capture identical instances across diverse viewpoints, thus providing authentic textures for cross-view transfer.
Building on this insight, we design an automated VLM-driven curation pipeline and construct \emph{TexRefSR-141K}, the first large-scale dataset tailored for identical-instance texture-reference SR.
As illustrated in Fig.~\ref{fig:data_pipeline}, the pipeline first assembles identical-instance triplets from the multi-view gallery, and then annotates them with a unified caption and complementary mask pairs via our semantic and spatial condition extraction.

\paragraph{Identical-Instance Triplet.} 
Departing from conventional reference paradigms, we retrieve triplets that depict strictly identical instances from multi-view galleries, guaranteeing genuine texture correspondence rather than mere semantic resemblance.
Since faithful restoration demands a texture-rich reference, we deliberately anchor the retrieval process on the reference image.
Specifically, a quality filter selects an optimal HQ product image to serve as $I_{REF}$.
Guided by this anchor, the pipeline retrieves a relevant HQ view of the identical instance to act as the ground-truth $I_{HQ}$.
A high-order degradation model~\cite{wang2021real} is then applied to $I_{HQ}$ to synthesize the LQ input $I_{LQ}$, formulating the core $\{I_{REF}, I_{HQ}, I_{LQ}\}$ triplet.
During practical inference, $I_{LQ}$ originates directly from real-world observations, and its corresponding $I_{REF}$ is retrieved from the instance gallery.

\paragraph{Caption and Complementary Masks.} 
Beyond mere image pairing, we also equip each triplet with explicit spatial-semantic guidance to resolve the critical ambiguities of \emph{what} and \emph{where} to transfer.
Leveraging the same semantic and spatial condition extraction introduced in our architecture, we obtain the unified caption $T$ and the complementary mask pair $\{M_{REF}, M_{TGT}\}$ for each triplet.
These masks cleanly isolate valid texture regions from irrelevant backgrounds and precisely localize the target instance across unaligned views.
To maximize textual supervision quality during training, we derives $T$ and $M_{TGT}$ utilizing the clearer $I_{HQ}$.
Additionally, random morphological dilation is applied to both spatial masks to ensure robustness along delicate boundaries (\textit{e.g.}, lace and decorative edges).

\begin{figure}[!t]
    \centering
    \includegraphics[width=1.0\linewidth]{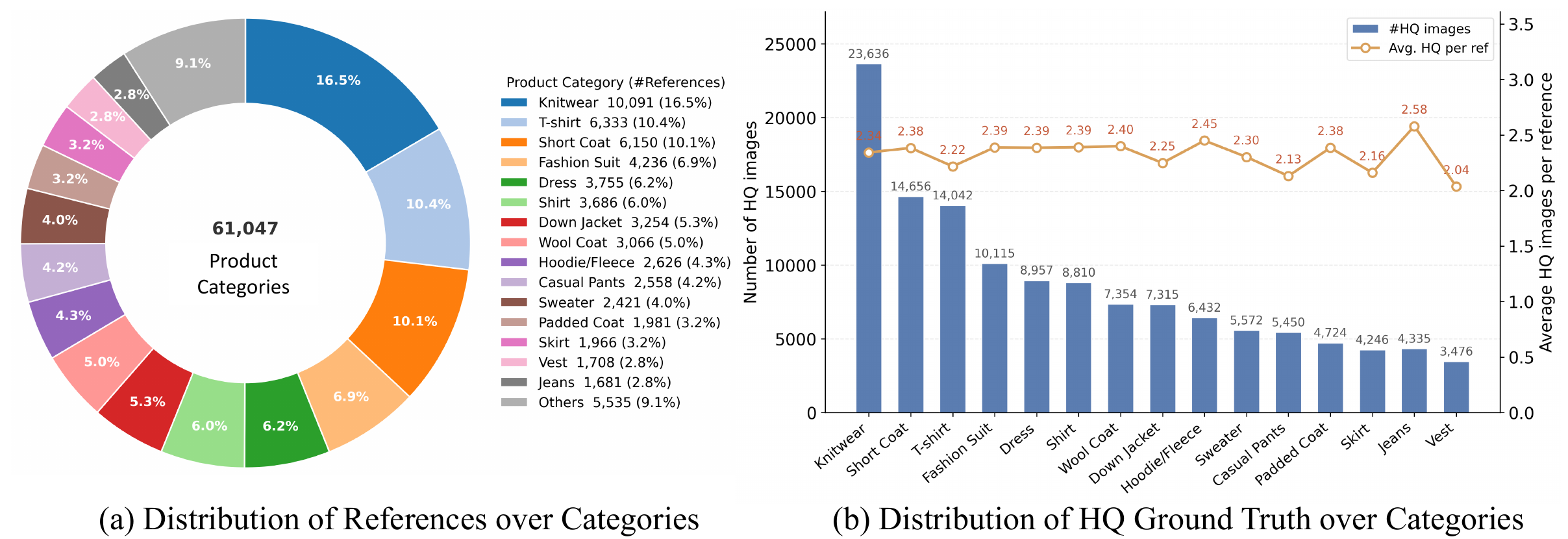}
    \caption{Dataset statistics. (a) Reference distribution over the top-15 leaf categories, with long-tail categories merged as "Others". (b) Number of HQ images per category (bars) and the average number of HQ images per reference (line).}
  \label{data_distribution}
\end{figure}

\paragraph{Quality Assessment and Dataset Statistics.}
To guarantee the reliability of reference guidance, an initial filtering stage strictly assesses candidate images before masking: it computes embedding similarities to discard mismatched pairs, eliminates degraded samples via a quality predictor, and rejects extreme close-ups that lose recognizable instance identity.
Through this rigorous pipeline, we construct TexRefSR-141K, comprising over 141K high-quality tuples across 61K diverse fashion products (Fig.~\ref{data_distribution}~(a)).
Crucially, each reference maps to multiple target ground-truths (Fig.~\ref{data_distribution}~(b)).
This $1$-to-$N$ identical-instance mapping yields rich cross-view texture correspondences, endowing the network with the spatial selectivity essential for authentic texture-reference-guided SR.

Beyond the training data, we further curate \emph{TexRefSR-Eval}, the first benchmark dedicated to identical-instance reference-guided SR, comprising 300 
manually verified test cases. It consists of two complementary subsets: \emph{TexRefSR-Eval-Syn} (250 samples), whose LQ inputs are synthesized from high-quality ground truths for paired full-reference evaluation, and \emph{TexRefSR-Eval-Real} (50 samples), whose LQ inputs are naturally degraded images collected from raw e-commerce galleries for real-world generalization. Together, they form a complete foundation for training and benchmarking this task.

\input{table}

\section{Experiments}

\subsection{Experimental Setup}

\paragraph{Benchmarks.}
We evaluate reference-guided real-world image SR on our newly established 
\textit{TexRefSR-Eval} benchmark, whose scale factor follows each subset's 
degradation ($4\times$ for the \textit{Syn} subset and native resolution for the \textit{Real} subset). To further assess general SR performance, we adopt two standard real-world datasets, 
RealSR~\cite{cai2019toward} and DRealSR~\cite{wei2020component}, both under 
$4\times$ upscaling following OSEDiff~\cite{wu2024one} for fair comparison.


\paragraph{Baselines.}
We compare against methods from three categories: (i) \textit{Generative SR}, including multi-step methods SUPIR~\cite{yu2024scaling} and DiT4SR~\cite{duan2025dit4sr}, and one-step methods OSEDiff~\cite{wu2024one}, PiSA-SR~\cite{sun2025pixel}, ODTSR~\cite{fang2026one}, and VOSR~\cite{wu2026vosr}; (ii) \textit{Reference-based SR}, including RefSR~\cite{guo2024refir}, AdaRefSR~\cite{wang2026trust}, and GarmentZoom~\cite{zhao2026garmentzoom}; and (iii) \textit{Commercial editing models}, including Gemini-3-Pro and GPT-Image-2. For open-source methods, we obtain all results using their official codes and models under default settings, while closed-source commercial models are evaluated through their official APIs.

\paragraph{Evaluation Metrics.}
For evaluation, we adopt both full-reference (FR) and no-reference (NR) metrics. The FR metrics consist of the pixel-level metrics PSNR and SSIM~\cite{wang2004image} and the perceptual-level metrics LPIPS~\cite{zhang2018lpips} and DISTS~\cite{ding2020dists}. For NR image quality assessment, we report the results of NIQE~\cite{mittal2012making}, MUSIQ~\cite{ke2021musiq}, CLIPIQA~\cite{wang2023exploring}, and MANIQA~\cite{yang2022maniqa}. In addition, to evaluate real-world reference-guided restoration, we introduce two VLM-based measures alongside their average (Overall): perceptual quality (PQ) of the predicted images to assess the overall visual realism of the restoration, and texture consistency (TC) computed within the masked regions between the reference and predicted images to evaluate their texture fidelity.

\begin{figure*}[!t]
    \centering
    \includegraphics[width=0.96\linewidth]{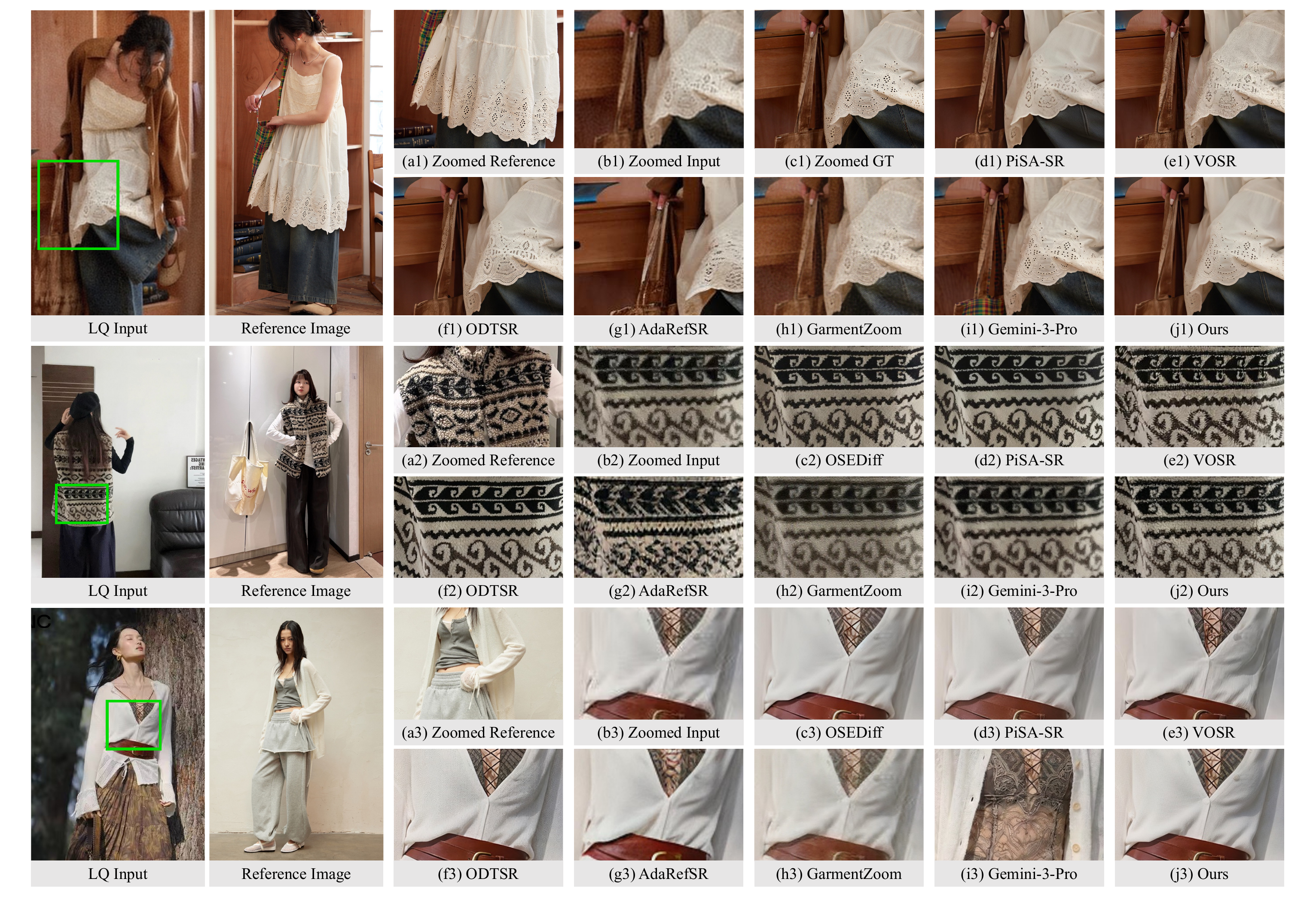}
    \caption{Qualitative comparison on TexRefSR-Eval, including one synthetic case with GT (top) and two real cases (bottom). Our method achieves reference-faithful restoration with superior fidelity and texture consistency. (\textbf{Zoom in for best view})
}
  \label{fig:qualitative}
\end{figure*}

\subsection{Main Results on Texture-Guided Real-World SR}

\label{sec:quant}
Quantitative comparisons on the TexRefSR-Eval benchmarks are reported in Tables~\ref{tab:texrefsr_syn} and~\ref{tab:texrefsr_real}. Based on these results, we draw the following conclusions: \textbf{\textit{(i)} Our method achieves the optimal balance between reference fidelity and perceptual quality.} As shown in Table~\ref{tab:texrefsr_syn}, GraftSR dominates all full-reference metrics. Specifically, it reduces LPIPS by 20.2\% compared to the second-best method (ODTSR) and is the only approach to achieve a DISTS score strictly below 0.10. This physical fidelity is accomplished without compromising visual aesthetics, as evidenced by securing the highest CLIPIQA (0.7275) and highly competitive MUSIQ scores. \textbf{\textit{(ii)} Our method yields the best texture consistency in real-world e-commerce scenarios.} On TexRefSR-Eval-Real (Table~\ref{tab:texrefsr_real}), GraftSR leads the VLM evaluation, which assesses semantic texture alignment. It achieves the highest texture consistency (2.840, compared to 2.700 for Gemini-3-Pro and 2.380 for VOSR), perceptual quality (3.980), and overall score (3.410). \textbf{\textit{(iii)} Our method outperforms powerful closed-source commercial models in texture-guided real-world SR tasks.} Compared to GPT-Image-2 and Gemini-3-Pro, our method demonstrates significantly higher fidelity on the synthetic benchmark (PSNR of 30.11\,dB vs.\ 18.33\,dB and 26.79\,dB) and a superior VLM overall score on the real-world benchmark (3.410 vs.\ 2.560 and 3.290). This gap arises because commercial models tend to over-edit and introduce hallucinated details, whereas GraftSR strictly preserves the pixel-level textures critical for reliable restoration.

\input{table_general}

-quality texture-reference-guided image super-resolution.
\begin{figure*}[!t]
    \centering
    \includegraphics[width=0.9\linewidth]{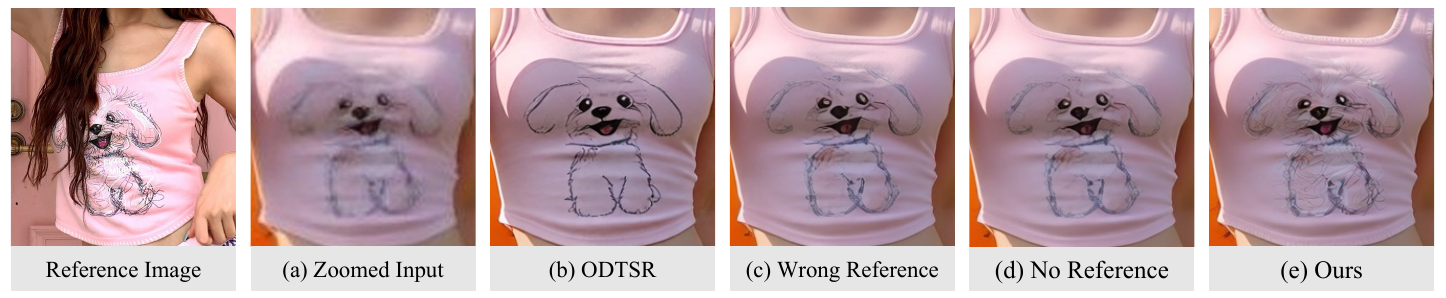}
    \caption{Robustness analysis of GraftSR with absent or mismatched references during inference.}
  \label{fig:robustness_ablation}
\end{figure*}

\subsection{Main Results on General Real-World SR}
Table~\ref{tab:general_real_isr} reports the results on the RealSR and DRealSR benchmarks where the reference branch is left empty. This setup explicitly evaluates whether our framework maintains its general restoration capability without reference guidance. Notably, GraftSR achieves the highest fidelity on both benchmarks, outperforming the second-best PiSA-SR by a significant margin on DRealSR (+1.09,dB PSNR and +0.029 SSIM). Furthermore, GraftSR maintains a solid balance between fidelity and perception, ranking second in LPIPS while securing competitive no-reference scores (MUSIQ and MANIQA). These results demonstrate the strong robustness of our GraftSR when references are unavailable. We attribute this to injecting reference cues via sequence concatenation, which makes the reference branch a detachable component that gracefully disengages when references are absent, allowing the framework to subsume general real-world SR.

\begin{figure}[!b]
    \centering
    \includegraphics[width=0.92\linewidth]{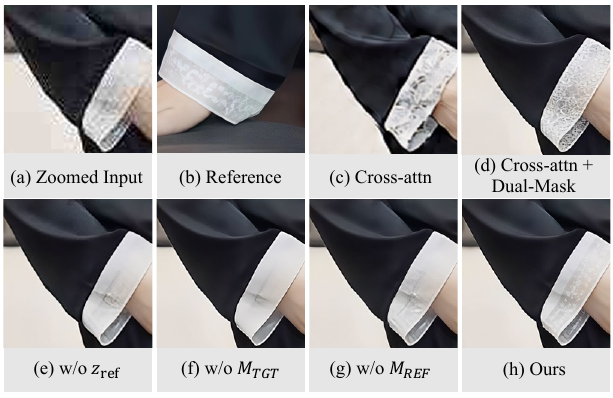}
    \caption{Visual comparison of the ablation studies.}
  \label{fig:ablation_visual}
\end{figure}

\input{table_ablation_merged}

\subsection{Qualitative Comparison}

Figure~\ref{fig:qualitative} presents a qualitative comparison on TexRefSR-Eval, including one synthetic case with ground truth~(GT) and two real cases. On the synthetic case, our result attains both high fidelity to the GT and strong texture consistency with the reference on the delicate lace~(j1), whereas Gemini-3-Pro alters the background bag, degrading fidelity to the GT~(i1). On real cases, blind generative SR methods (\textit{e.g.}, ODTSR) hallucinate textures inconsistent with the reference~(f2--f3), while reference-based SR methods (\textit{e.g.}, AdaRefSR, GarmentZoom) fail beyond matched regions and produce artifacts due to brittle spatial misalignment~(g2--g3, h2--h3). In contrast, only GraftSR faithfully recovers the fluffy sherpa texture~(j2) and the fine knit with correct buttons~(j3), demonstrating its robustness in texture-reference-guided SR.

\subsection{In-depth Analysis}
\label{sec:analysis}

\noindent\textbf{Texture Reference Injection.}~
Table~\ref{tab:ablation_merged} (top) and Fig.~\ref{fig:ablation_visual}~(c--d) evaluate injection strategies. \textit{Cross-attn} lacks texture consistency despite mask guidance. Unmasked \textit{Joint-attn} improves texture transfer but introduces structural artifacts (higher NIQE). Ultimately, our full model synergizes joint attention for maximal transfer and dual masks for precise alignment.

\noindent\textbf{Dual-Mask Reference Guidance.}~
As shown in Table~\ref{tab:ablation_merged} (bottom) and Fig.~\ref{fig:ablation_visual}~(e--g), removing the reference token $z_{\text{ref}}$ degrades texture consistency, failing to synthesize lace textures~(Fig.~\ref{fig:ablation_visual}~(e)). In addition, discarding either spatial mask ($M_{TGT}$ or $M_{REF}$) disrupts the reference-target alignment, causing textures to nearly vanish~(Fig.~\ref{fig:ablation_visual}~(f--g)). In contrast, full GraftSR accurately localizes delicate textures~(Fig.~\ref{fig:ablation_visual}~(h)), confirming all components are indispensable.

\begin{figure}[!t]
    \centering
    \includegraphics[width=0.9\linewidth]{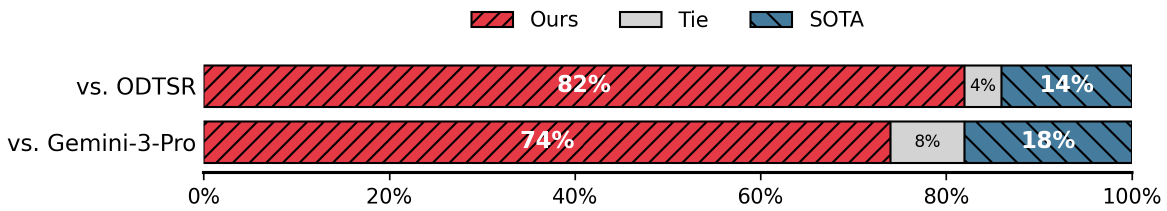}
    \caption{The user study results. Our method is consistently preferred over SOTA ODTSR and Gemini-3-Pro.}
  \label{fig:user_study}
\end{figure}

\noindent\textbf{User Study.} We conduct a Two-Alternative Forced Choice (2AFC) study across 50 diverse scenes, where participants judge our results against the strongest SOTA methods based on perceptual quality and texture consistency. As shown in Fig.~\ref{fig:user_study}, our method wins \textbf{82\%} of votes against ODTSR and \textbf{74\%} against the leading Gemini-3-Pro. This demonstrates a strong alignment with human visual preferences.

\noindent\textbf{Robustness Analysis and Limitations.} As shown in Fig.~\ref{fig:robustness_ablation}, we test GraftSR with a mismatched reference or no reference during inference. In both cases, it still produces faithful textures comparable to using the correct reference and clearly outperforms ODTSR. 
Nevertheless, when inputs are severely degraded, GraftSR trades off some fine-grained details to preserve fidelity, which we leave for future work.

\section{Conclusion}
In this paper, we present GraftSR, a texture-reference-guided generative SR framework that restores authentic textures from identical-instance references to overcome the texture hallucination of diffusion-based real-world SR. To handle the inherent misalignment between low-quality inputs and their references, GraftSR introduces a dual-mask reference guidance mechanism that decouples \emph{what} textures to extract from \emph{where} to apply them, enabling robust transfer without brittle spatial alignment. We further construct TexRefSR-141K and TexRefSR-Eval, the first large-scale identical-instance dataset and benchmark with complementary spatial masks. Extensive experiments show that GraftSR sets a new state-of-the-art for highly faithful texture restoration.

\bibliography{references}
\input{appendix}

\end{document}

%% file: table.tex
\definecolor{bestred}{RGB}{204,0,0}
\definecolor{secondblue}{RGB}{0,0,204}
\newcommand{\best}[1]{\textbf{\textcolor{bestred}{#1}}}
\newcommand{\second}[1]{\underline{\textcolor{secondblue}{#1}}}

\begin{table*}[!t]
\centering
\resizebox{\textwidth}{!}{
\begin{tabular}{ll cccc cccc}
\toprule
\multirow{2}{*}{Type} & \multirow{2}{*}{Method}
& \multicolumn{4}{c}{Full-Reference} & \multicolumn{4}{c}{No-Reference} \\
\cmidrule(lr){3-6} \cmidrule(lr){7-10}
& & LPIPS$\downarrow$ & DISTS$\downarrow$ & PSNR$\uparrow$ & SSIM$\uparrow$
& MUSIQ$\uparrow$ & NIQE$\downarrow$ & CLIPIQA$\uparrow$ & MANIQA$\uparrow$ \\
\midrule
\multirow{6}{*}{\shortstack[l]{Open-source\\Generative SR}}
& OSEDiff   & \second{0.2326} & 0.1444 & \second{28.3752} & \second{0.8471} & 65.2201 & 4.8256 & 0.6631 & 0.5250 \\
& SUPIR     & 0.2208 & 0.1321 & 29.3151 & 0.8382 & 61.5577 & 5.2317 & 0.5922 & 0.5225 \\
& DiT4SR    & 0.3816 & 0.2147 & 25.6121 & 0.8071 & 33.1845 & 7.3649 & 0.5769 & 0.4253 \\
& ODTSR     & 0.1983 & \second{0.1048} & 29.5098 & 0.8431 & 69.2601 & \second{4.7789} & 0.6176 & 0.5129 \\
& PiSA-SR   & 0.2247 & 0.1343 & 28.5478 & 0.8458 & 66.7754 & \best{4.6765} & \second{0.6831} & 0.5338 \\
& VOSR-1.4b & 0.2349 & 0.1247 & 28.3661 & 0.8290 & \best{71.1585} & 4.3381 & 0.6873 & \best{0.5523} \\
\midrule
\multirow{2}{*}{\shortstack[l]{Open-source\\Reference-based SR}}
& RefSR     & 0.4384 & 0.2685 & 28.6190 & 0.8297 & 24.0680 & 8.1108 & 0.4106 & 0.3259 \\
& AdaRefSR  & 0.6071 & 0.2785 & 16.2155 & 0.7154 & 32.1215 & 7.7052 & 0.5645 & 0.4367 \\
& GarmentZoom  & 0.4156 & 0.2643 & 25.8811 & 0.7255 & 34.1616 & 6.1379 & 0.4360 & 0.4160 \\
\midrule
\multirow{2}{*}{\shortstack[l]{Commercial\\Editing Models}}
& GPT-Image-2  & 0.4474 & 0.2154 & 18.3255 & 0.7665 & 34.4500 & 7.4755 & 0.5781 & 0.4355 \\
& Gemini-3-Pro & 0.2065 & 0.1034 & 26.7874 & 0.8222 & 64.9845 & 4.9920 & 0.6216 & 0.5048 \\
\midrule
& \textbf{Graft~(Ours)} & \best{0.1583} & \best{0.0986} & \best{30.1060} & \best{0.8494}
& \second{69.8535} & 4.9346 & \best{0.7275} & \second{0.5270} \\
\bottomrule
\end{tabular}}
\caption{Quantitative comparison on the \textbf{TexRefSR-Eval-Syn} benchmark, where ground-truth references are available. Both full-reference and no-reference metrics are reported. \best{Red bold} and \second{blue underline} denote the best and the second-best results.}
\label{tab:texrefsr_syn}
\end{table*}

\begin{table*}[!t]
\centering
\resizebox{\textwidth}{!}{
\begin{tabular}{ll cccc ccc}
\toprule
\multirow{2}{*}{Type} & \multirow{2}{*}{Method}
& \multicolumn{4}{c}{No-Reference}
& \multicolumn{3}{c}{VLM Evaluation (GPT-4.1-global)} \\
\cmidrule(lr){3-6} \cmidrule(lr){7-9}
& & NIQE$\downarrow$ & MUSIQ$\uparrow$ & CLIPIQA$\uparrow$ & MANIQA$\uparrow$
& PQ$\uparrow$ & TC$\uparrow$ & Overall$\uparrow$ \\
\midrule
\multirow{6}{*}{\shortstack[l]{Open-source\\Generative SR}}
& OSEDiff    & 3.9329 & 70.5668 & 0.6395 & 0.6292 & 3.320 & 1.740 & 2.530 \\
& SUPIR      & 4.2105 & 59.6691 & 0.4778 & 0.5741 & 3.380     & 1.960     & 2.670     \\
& DiT4SR     & \best{3.7190} & 72.5782 & \best{0.6428} & 0.6485 & 3.720 & 2.040 & 2.880 \\
& ODTSR      & 4.2809 & \second{71.5261} & 0.6140 & \best{0.6580} & 2.760 & 1.560 & 2.160 \\
& PiSA-SR    & 4.2523 & 65.5855 & 0.4913 & 0.5789 & 2.840 & 1.660 & 2.250 \\
& VOSR-1.4b  & \second{3.8273} & 69.7636 & 0.6227 & 0.6399 & \second{3.880} & 2.380 & 3.130 \\
\midrule
\multirow{2}{*}{\shortstack[l]{Open-source\\Reference-based SR}}
& RefSR      & 5.0782 & 53.8617 & 0.3251 & 0.5030 & 2.200 & 1.280 & 1.740 \\
& AdaRefSR   & 5.7189 & 46.2674 & 0.3625 & 0.5087 & 2.580 & 1.560 & 2.070 \\
& GarmentZoom   & 5.6298 & 31.4101 & 0.3625 & 0.4417 & 2.380 & 1.840 & 2.110 \\
\midrule
\multirow{2}{*}{\shortstack[l]{Commercial\\Editing Models}}
& GPT-Image-2   & 4.9825 & 65.9262 & 0.5667 & 0.6159 & 3.020 & 2.100 & 2.560 \\
& Gemini-3-Pro  & 5.1077 & 56.5937 & 0.4633 & 0.5277 & \second{3.880} & \second{2.700} & \second{3.290} \\
\midrule
& \textbf{GraftSR~(Ours)} & 3.9881 & \best{73.0045} & \second{0.6409} & \second{0.6507}
& \best{3.980} & \best{2.840} & \best{3.410} \\
\bottomrule
\end{tabular}}
\caption{Quantitative comparison on the \textbf{TexRefSR-Eval-Real} benchmark with no-reference metrics and VLM-based metrics.}
\label{tab:texrefsr_real}
\end{table*}

%% file: table_general.tex
\begin{table}[!b]
\centering
\resizebox{\columnwidth}{!}{
\begin{tabular}{c l ccccc}
\toprule
\textbf{Dataset} & \textbf{Method} 
& PSNR$\uparrow$ & SSIM$\uparrow$ & LPIPS$\downarrow$
& MUSIQ$\uparrow$ & MANIQA$\uparrow$ \\
\midrule
\multirow{7}{*}{\rotatebox{90}{RealSR}} 
& SUPIR & 23.65 & 0.6620 & 0.3541 & 62.09 & 0.5780 \\
& DiT4SR & 23.50 & 0.6657 & 0.3215 & 67.76 & 0.6564 \\
& OSEDiff & 25.15 & 0.7341 & 0.2920 & 69.08 & 0.6335 \\
& PiSA-SR & \second{25.40} & \second{0.7418} & 0.2672 & 70.14 & 0.6551 \\
& VOSR-1.4B & 25.23 & 0.7175 & 0.2732 & \best{70.58} & 0.6443 \\
& ODTSR & 25.07 & 0.7361 & \best{0.2398} & 68.29 & \second{0.6622} \\
& \textbf{GraftSR~(Ours)} & \best{25.51} & \best{0.7512} & \second{0.2559} & \second{70.37} & \best{0.6676} \\
\midrule
\multirow{7}{*}{\rotatebox{90}{DRealSR}} 
& SUPIR & 25.09 & 0.6460 & 0.4243 & 58.79 & 0.5471 \\
& DiT4SR & 25.40 & 0.6657 & 0.3869 & 65.75 & \second{0.6287} \\
& OSEDiff & 27.92 & \second{0.7835} & 0.2968 & 64.65 & 0.5895 \\
& PiSA-SR & \second{28.32} & 0.7804 & 0.2960 & \second{66.11} & 0.6146 \\
& VOSR-1.4B & 27.88 & 0.7413 & 0.3260 & 66.04 & 0.6053 \\
& ODTSR & 28.14 & 0.7736 & \best{0.2592} & 62.86 & 0.6227 \\
& \textbf{GraftSR~(Ours)} & \best{29.04} & \best{0.8052} & \second{0.2858} & \best{66.73} & \best{0.6376} \\
\bottomrule
\end{tabular}}
\caption{
Quantitative comparison on geneal real-world SR. our GraftSR still achieves strong generalization in both fidelity and perceptual quality even without reference images.
}
\label{tab:general_real_isr}
\end{table}

%% file: table_ablation_merged.tex
\begin{table}[!t] 
\centering
\setlength{\tabcolsep}{4pt}
\resizebox{\columnwidth}{!}{
\begin{tabular}{l cccc cc}
\toprule
Variant
& NIQE$\downarrow$
& MUSIQ$\uparrow$
& CLIPIQA$\uparrow$
& MANIQA$\uparrow$
& PQ$\uparrow$
& TC$\uparrow$ \\
\midrule
\multicolumn{7}{c}{\textit{Ablation on Texture Reference Injection}} \\
\midrule
Cross-attn
& 4.2210 & 70.55 & 0.5236 & 0.5874 & 2.660 & 1.840 \\
Cross-attn + Dual-Mask
& \best{3.9818} & 71.83 & 0.5974 & 0.6168 & 3.000 & 2.000 \\
Joint-attn
& 4.4200 & 71.07 & 0.5835 & 0.6278 & 3.400 & 2.100 \\
\midrule
\multicolumn{7}{c}{\textit{Ablation on Dual-Mask Reference Guidance}} \\
\midrule
w/o $z_{\text{ref}}$
& 4.2911 & \second{72.60} & 0.6222 & \second{0.6469} & 3.500 & 2.220 \\
w/o $M_{TGT}$
& 4.6184 & 71.25 & 0.5688 & 0.6279 & 3.500 & 2.360 \\
w/o $M_{REF}$
& 4.4125 & 72.23 & 0.5959 & 0.6375 & \second{3.640} & 2.440 \\
\midrule
\textbf{GraftSR~(Ours)}
& \second{3.9881} & \best{73.00} & \best{0.6409} & \best{0.6507} & \best{3.980} & \best{2.840} \\
\bottomrule
\end{tabular}}
\caption{Ablation studies on the texture reference injection strategy and the dual-mask reference guidance module.}
\label{tab:ablation_merged}
\end{table}

%% file: appendix.tex
\makeatletter
\newcommand{\maketitlesupplementary}[1]{%
  \twocolumn[{%
    \vbox{%
      \hsize\textwidth
      \linewidth\textwidth
      \vskip 0.625in minus 0.125in
      \centering
      {\LARGE\bfseries #1\par}
      \vskip 1em
    }%
  }]%
}
\makeatother

\setcounter{page}{1}
\clearpage
\maketitlesupplementary{%
  GraftSR: Grafting Authentic Textures for Real-World Image Super-Resolution\\
  via Identical-Instance Guidance\\[0.5em]
  \Large Supplementary Material
}

\setcounter{equation}{0}
\setcounter{figure}{0}

\section{Overview}
In this supplementary material, we present:

\begin{itemize}
	\item \textbf{Method details}, including the architecture details and the data construction procedure.
	
	\item \textbf{Implementation details}, including 
	hyperparameter settings, prompts, and training strategies.

	\item \textbf{User study details}.
	
	\item \textbf{More visualization results} on each dataset.
	
\end{itemize}

\appendix
\section{Method Details}
\label{sec:method}

In this section, we elaborate on the training details omitted from the main paper, including the one-step training formulation, the detailed training objectives, and the detail of our identical-instance dataset construction.

\subsection{One-Step Training Formulation}

\paragraph{Dual-Noise Latent Construction.}
As described in the main paper, we adopt a dual-noise strategy~\cite{fang2026one} to balance perception and fidelity. Given the low-quality latent $z_{LQ}=\mathcal{E}_{\text{VAE}}(I_{LQ})$, both the perception and control branches follow the flow-matching forward process:
\begin{equation}
\begin{aligned}
z &= (1-\sigma_{p})z_{LQ} + \sigma_{p}\epsilon, \ \sigma_{p} = \sigma(s_p), \\
z_{c}  &= (1-\sigma_{c})z_{LQ} + \sigma_{c}\epsilon, \ \sigma_{c} = \sigma(s_c),
\end{aligned}
\quad \epsilon\sim\mathcal{N}(0,\mathbf{I}),
\end{equation}
where $\sigma_p$ and $\sigma_c$ are the noise levels at timesteps $s_p{=}750$ and
$s_c\sim\mathcal{U}(750,999)$. A larger sampled $s_c$ yields weaker noise,
allowing the control branch to preserve fine details.

\paragraph{One-Step Velocity Prediction.}
To convert the multi-step denoising into a single forward pass, we adopt one-step adversarial training~\cite{lin2025diffusion} and skip the consistency-distillation stage, since our input is a noised low-quality latent rather than pure noise. The
generator predicts the velocity field to update the prior-noise latent to the high-quality target in a single step:
\begin{equation}
    z_{\text{pred}} = z + (0 - \sigma_p)\,v_\theta(z, z_{\text{c}}, z_{\text{sem}}, z_{\text{ref}}),
\end{equation}
which is decoded to $I_{\text{pred}} = D(z_{\text{pred}})$ for supervision. Note that when GPU memory is limited, high-resolution inputs can be processed in a tiled manner. In this case, in addition to the dual-noise encoding of each cropped patch, we encode the full image with the VAE and concatenate its latent as an additional global structural token into the control branch, so that each patch retains access to the global context. This extends the conditioning input of $v_\theta$ with one extra token, while the velocity-prediction and update rule remain unchanged.

\subsection{Training Objectives}

\paragraph{Generator Objective.}
The generator is trained with a reconstruction loss $\mathcal{L}_{\text{rec}}$ in RGB space and an adversarial loss $\mathcal{L}_{\text{adv}}^{G}$ in latent space. Beyond the standard MSE and LPIPS terms, we introduce the DISTS loss~\cite{ding2020dists}. Unlike LPIPS, which strictly penalizes spatial shifts, DISTS tolerates mild misalignment, making it well suited to our cross-view texture restoration. It extracts multi-stage feature maps from a pre-trained VGG network and jointly measures structural and textural similarity:
\begin{equation}
    \mathcal{L}_{\text{DISTS}} = 1 - \sum_{i=0}^{n} \left( \alpha_i \, l_s(\hat{F}_i, F_i) + \beta_i \, l_t(\hat{F}_i, F_i) \right),
\end{equation}
where $\hat{F}_i$ and $F_i$ are the feature maps of $I_{\text{pred}}$ and $I_{HQ}$ at the $i$-th VGG layer of a pre-trained VGG16 network($i=0$ denotes the input image and $n=5$ corresponds to the five convolutional stages), $l_s$ and $l_t$ measure structural and textural similarity, and $\alpha_i, \beta_i$ are the pre-trained weights from~\cite{ding2020dists}. This encourages texture fidelity without enforcing rigid pixel-to-pixel matching. The overall reconstruction loss is then formulated as:
\begin{equation}
    \mathcal{L}_{\text{rec}} = \mathcal{L}_{\text{MSE}}(I_{\text{pred}}, I_{HQ}) + \lambda_{\text{lpips}}\mathcal{L}_{\text{LPIPS}} + \lambda_{\text{dists}}\mathcal{L}_{\text{DISTS}}.
\end{equation}
For the adversarial term, we adopt a relativistic GAN loss~\cite{jolicoeur2018relativistic} to avoid mode collapse:
\begin{equation}
    \mathcal{L}_{\text{adv}}^{G} = -\mathbb{E}[\log(1 - R(z_{HQ}, z_{\text{pred}}))] - \mathbb{E}[\log(R(z_{\text{pred}}, z_{HQ}))],
\end{equation}
where $z_{HQ} = \mathcal{E}_{\text{VAE}}(I_{HQ})$ is the clean latent encoded from the high-quality target image, and $R(a, b) = \mathcal{S}(D(a) - D(b))$ measures
the relativistic realism between two latents. 
Since the VAE encoder $\mathcal{E}_{\text{VAE}}^{'}$ in the prior noise branch is trainable, we further introduce a VAE loss to prevent it from drifting away from the target distribution. Specifically, its encoded LQ latent $z_{LQ}$ is decoded by the frozen VAE decoder and supervised against the HQ target:
\begin{equation}
\mathcal{L}_{\text{vae}} = \mathcal{L}_{\text{MSE}}(D\left(\mathcal{E}_{\text{VAE}}'(I_{LQ})\right) - I_{HQ} )
\end{equation}
The overall generator objective is:
\begin{equation}
\mathcal{L}_G = \mathcal{L}_{\text{rec}} + \lambda_{\text{adv}}\mathcal{L}_{\text{adv}}^{G} + \lambda_{\text{vae}}\mathcal{L}_{\text{vae}}
\end{equation}

\paragraph{Discriminator Objective.}
Following the one-step SR paradigm~\cite{fang2026one}, our discriminator is built upon the Wan2.1 1.3B~\cite{wan2025wan} architecture, which shares the same VAE encoder as the Qwen-Image backbone and injects textual conditioning through cross-attention. Instead of introducing extra cross-attention layers to regress a single scalar as in as in Adversarial Post-Training (APT)~\cite{lin2025diffusion}, we simply append two 2D convolutional layers to the transformer outputs, yielding patch-wise realism scores. The discriminator is initialized from pre-trained weights and is optimized with an adversarial loss $\mathcal{L}_{\text{adv}}^{D}$ that is symmetric to the generator counterpart:
\begin{equation}
    \mathcal{L}_{\text{adv}}^{D} = -\mathbb{E}[\log(R(z_{HQ}, z_{\text{pred}}))] - \mathbb{E}[\log(1 - R(z_{\text{pred}}, z_{HQ}))].
\end{equation}
To stabilize training, we further impose an approximated R1 regularization that suppresses the discriminator gradients on real samples:
\begin{equation}
    \mathcal{L}_{R} = \left\| D(z_{HQ}) - D\!\left(z_{HQ} + \delta\right) \right\|_2^2,
    \quad \delta \sim \mathcal{N}(0, \sigma^2 \mathbf{I}),
\end{equation}
where $\delta$ is a small Gaussian perturbation added to the real latent $z_{HQ}$, encouraging consistent predictions between the clean and perturbed inputs. The
overall discriminator objective is:
\begin{equation}
    \mathcal{L}_D = \mathcal{L}_{\text{adv}}^{D} + \lambda_{R}\,\mathcal{L}_{R}.
\end{equation}

\subsection{Dataset Construction Details}
\label{sec:data}

We supplement two concrete details of our curation pipeline omitted from the
main paper: the mask dilation scheme and the pair quality assessment rules.

\paragraph{Mask Dilation.}
To avoid artifacts around delicate peripheral structures (\emph{e.g.}, lace,
tassels, decorative edges), we dilate the complementary masks with an elliptical
kernel followed by a light Gaussian smoothing. During training, the dilation
radius is randomly sampled per mask as a boundary-preserving augmentation, using
$3$--$6$ pixels for the target mask and a larger $5$--$10$ pixels for the
reference mask to tolerate cross-view misalignment; at inference, a fixed radius
is adopted for deterministic behavior.

\paragraph{Pair Quality Assessment.}
As mentioned in the main paper, an initial filtering stage strictly assesses candidate images before masking. We detail its three criteria here: (1)~we compute the semantic embedding similarity~\cite{zhang2025qwen3} between $I_{REF}$ and $I_{HQ}$ and discard any tuple below a strict threshold of $0.85$, which eliminates mismatched pairs such as a garment matched with an image showing only text-based specifications; (2)~we adopt the MDIQA~\cite{yao2025mdiqa} quality predictor to eliminate degraded samples with severe motion blur or compression artifacts, avoiding unreliable texture supervision; and (3)~we apply heuristic rules to reject extreme close-ups (\emph{e.g.}, macro shots of fabric) that lose recognizable instance identity and lack sufficient structural context.

\section{Implementation Details}
In this section, we provide the full implementation details of GraftSR to facilitate reproducibility. We first present the complete hyperparameter settings for both the generator and the discriminator. We then describe the training strategies adopted to enhance robustness and generalization. Finally, we detail the prompts used for VLM throughout the pipeline.
\label{sec:impl}

\subsection{Hyperparameter Settings}
We build the generator based on the Qwen-Image-Edit~\cite{wu2025qwen} (2511 version) backbone and fine-tune it with LoRA, while the Wan2.1-1.3B discriminator is fully fine-tuned from its pre-trained weights. The rank of LoRA is 128. Both networks are optimized using RMSprop~($\alpha{=}0.9$, momentum${=}0$) with constant learning rates of
$5\times10^{-5}$ for the generator and $5\times10^{-6}$ for the discriminator.
We train for one epoch with a total batch size of $32$ on $32$ NVIDIA H20 GPUs,
taking roughly $30$ hours, with gradient checkpointing enabled and the
discriminator updated once per generator step. Training is conducted on
$512\times512$ patches. The perception and control branches inject noise starting
from timesteps $s_p{=}750$ and $s_c{\sim}\mathcal{U}(750,999)$, respectively.
For the loss weights, we set the LPIPS, DISTS, and VAE weights to 
$\lambda_{\text{lpips}}{=}\lambda_{\text{dists}}{=}\lambda_{\text{vae}}{=}1.0$ with the MSE term weighted by $1.0$ by default, and the adversarial loss 
weight to $\lambda_{\text{adv}}{=}0.02$. The R1 regularization weight is set to 
$\lambda_{R}{=}5.0$ with a perturbation variance $\sigma^2{=}0.005$.

\subsection{Training Strategies}
To enhance robustness and generalization, we adopt several stochastic training strategies. To construct training patches, we randomly crop a $512\times512$
region, where with a probability of $0.5$ the crop is centered at a point sampled within the mask region to ensure the target instance is well covered, and otherwise a random region is selected. To prevent the model from over-relying on the auxiliary conditions, we randomly drop the entire reference branch and the spatial mask each with a probability of $0.1$, encouraging the network to remain effective even when the reference or mask is unavailable. In addition, with a probability of $0.5$ the control branch directly receives the clean low-quality latent without noise injection, which further balances detail preservation and restoration capability.

\subsection{VLM Prompts}
\label{sec:prompt}

\paragraph{Annotation Prompt.}
To jointly obtain the unified caption $T$ and the complementary masks, we prompt the VLM with the reference and target images that depict the same instance, asking it to identify the shared entity, describe its visual attributes, and localize the matched semantic region in both views for cross-view alignment:

\begin{tcolorbox}[colback=gray!5,colframe=gray!50,boxrule=0.5pt,breakable]
\small
\textbf{Task.} You will receive two images depicting the same product and its
associated outfit: a scene/model image (SRC) and a high-quality reference image
(REF). Localize the corresponding region of the same product in both images and
return the matched semantic bounding boxes for affine alignment.

\textbf{Key requirements.} (1)~\emph{Semantic consistency}: the two boxes must
enclose the same visible part of the same product; if SRC shows only the upper
part, the REF box should cover only the corresponding upper part rather than the
whole product. (2)~\emph{Completeness}: under semantic consistency, each box
should cover the product's visible region as completely as possible.

\textbf{Output.} \texttt{product\_entity} (the plain object noun, \emph{e.g.},
dress, T-shirt); \texttt{product\_description} (visual attributes such as color,
material, texture, and style); \texttt{src\_image\_description} (full visual
content of SRC); \texttt{match\_description}; and \texttt{src\_matched\_bbox} /
\texttt{ref\_matched\_bbox} (the matched region in each image).
\end{tcolorbox}

\paragraph{Evaluation Prompt.}
For the VLM-based evaluation on real-world SR, where no ground truth is available, we adopt a pointwise scoring protocol. The prediction is presented as a global view together with a full-resolution crop of the key region, and is graded on its intrinsic quality and its consistency with the reference crop. The VLM is instructed by the following system and task prompts, and rates each dimension on an integer scale of $1$--$5$:

\begin{tcolorbox}[colback=gray!5,colframe=gray!50,boxrule=0.5pt,breakable]
\small
\textbf{System.} You are a rigorous, objective image-quality evaluator for reference-based super-resolution. For each image you get a global view and a full-resolution crop of the key region: use the global view for content/composition and the crop for fine detail/texture. Return only a valid JSON object. Each score is an integer $1$--$5$ ($1$=worst, $5$=best), and the full range should be used.

\medskip
\textbf{Task.} Evaluate a real-world image restoration result. The inputs, each consisting of a global view and a key-region crop, are: (i)~\textbf{PRED}, the model output to evaluate; and (ii)~\textbf{REF}, the
reference showing the true textures of the key item. Grade PRED on its intrinsic quality and its consistency with REF along the dimensions below, and return the scores in the specified JSON format.

\medskip
\textbf{Evaluation dimensions} (the key goal is
restoring correct, faithful \emph{texture} of the key item):

\textbf{(1)~Perceptual quality:} overall visual quality --- sharpness, natural realistic details, correct tone/color, and absence of degradation. Penalize blur,
insufficient enhancement, noise, artifacts, over-smoothing, over-sharpening halos, over-saturation, and color shift. Note that real high-resolution images
naturally contain fine high-frequency textures; a visibly sharper yet faithful result should score \emph{higher}, and only \emph{false} details (halos, ringing, hallucinated textures) are penalized.

\textbf{(2)~Texture consistency:} whether the key item's textures/patterns match the REF, and whether they blend seamlessly in lighting, perspective, and
style. Penalize destroyed, fabricated, or mismatched textures.
\end{tcolorbox}

\begin{figure}[!t]
    \centering
    \includegraphics[width=\linewidth]{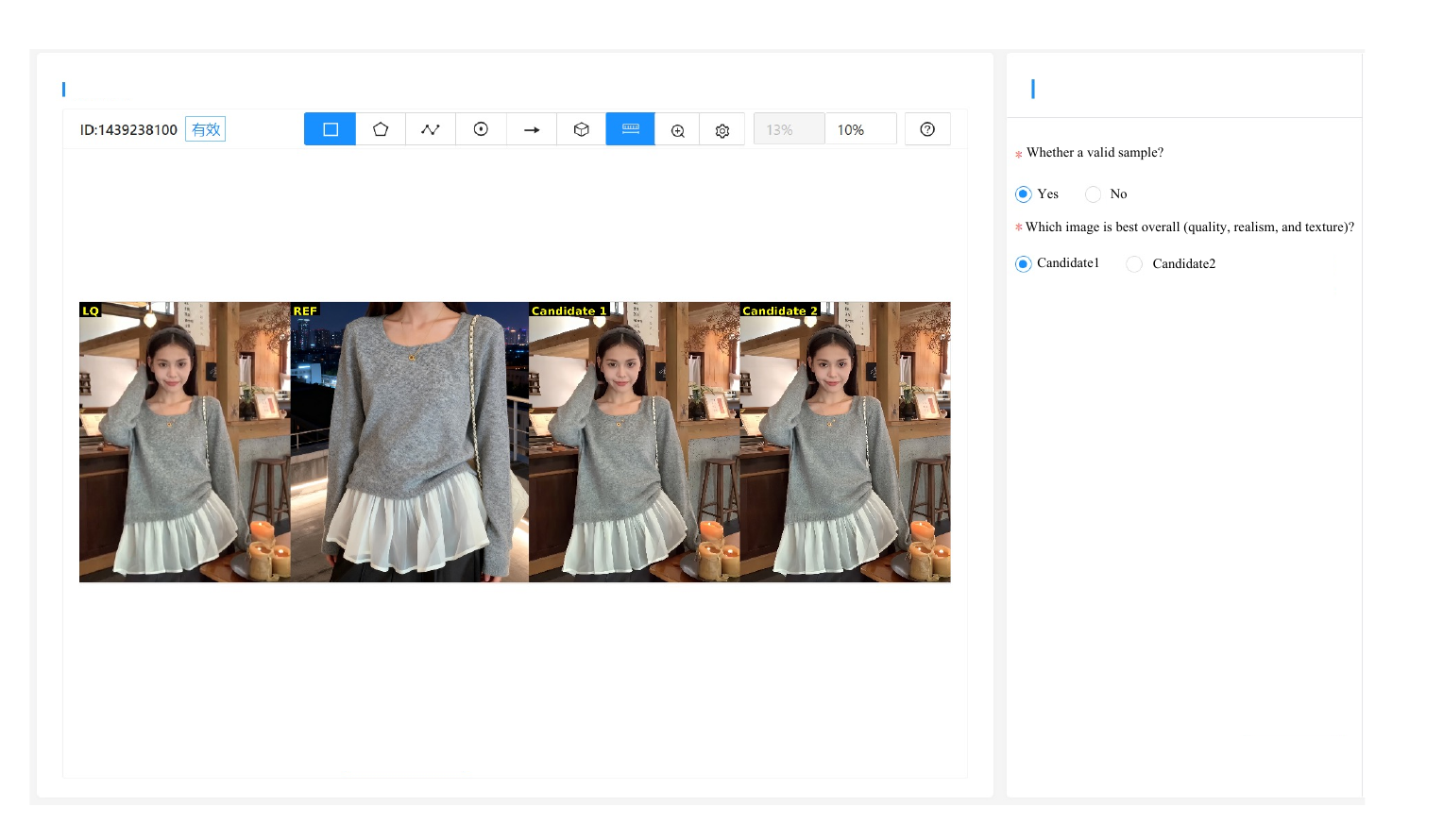}
    \caption{User Study Interface.}
    \label{fig:user_study_ui}
\end{figure}

\section{User Study Details} 

We conduct a two-alternative forced-choice (2AFC) study with 10 participants over 50 diverse scenes. In each trial, the low-quality input and the reference image are presented alongside two anonymized candidates, namely our result and that of a competing method, whose left and right positions are randomly shuffled across trials to eliminate positional bias. Participants are forced to select the better result according to perceptual quality and texture consistency with the reference.

To ensure consistent judgments, participants follow a set of evaluation criteria. First, regarding perceptual quality, each candidate is compared against the low-quality input, and a candidate is assigned a lower preference if it remains blurry or fails to recover basic details. Second, regarding texture consistency, each candidate is compared against the reference, focusing on whether the large-scale material and texture details within the product region are consistent, while highly localized regions are ignored. In addition, candidates are assigned a lower preference if their textures are inconsistent with the reference or their appearances are over-smoothed and deviate from the authentic material. Overall, sharpness is prioritized, such that a candidate without effective quality enhancement is not preferred, and among candidates with comparable sharpness the one with better texture preservation is selected. Furthermore, we exclude the option of whether invalid, such as those with an unusable reference or corrupted images, prior to aggregation. The user study interface is shown in Fig.~\ref{fig:user_study_ui}.

\section{More Visualization Results}
\label{sec:vis}

\begin{figure*}[!t]
    \centering
    \includegraphics[width=\linewidth]{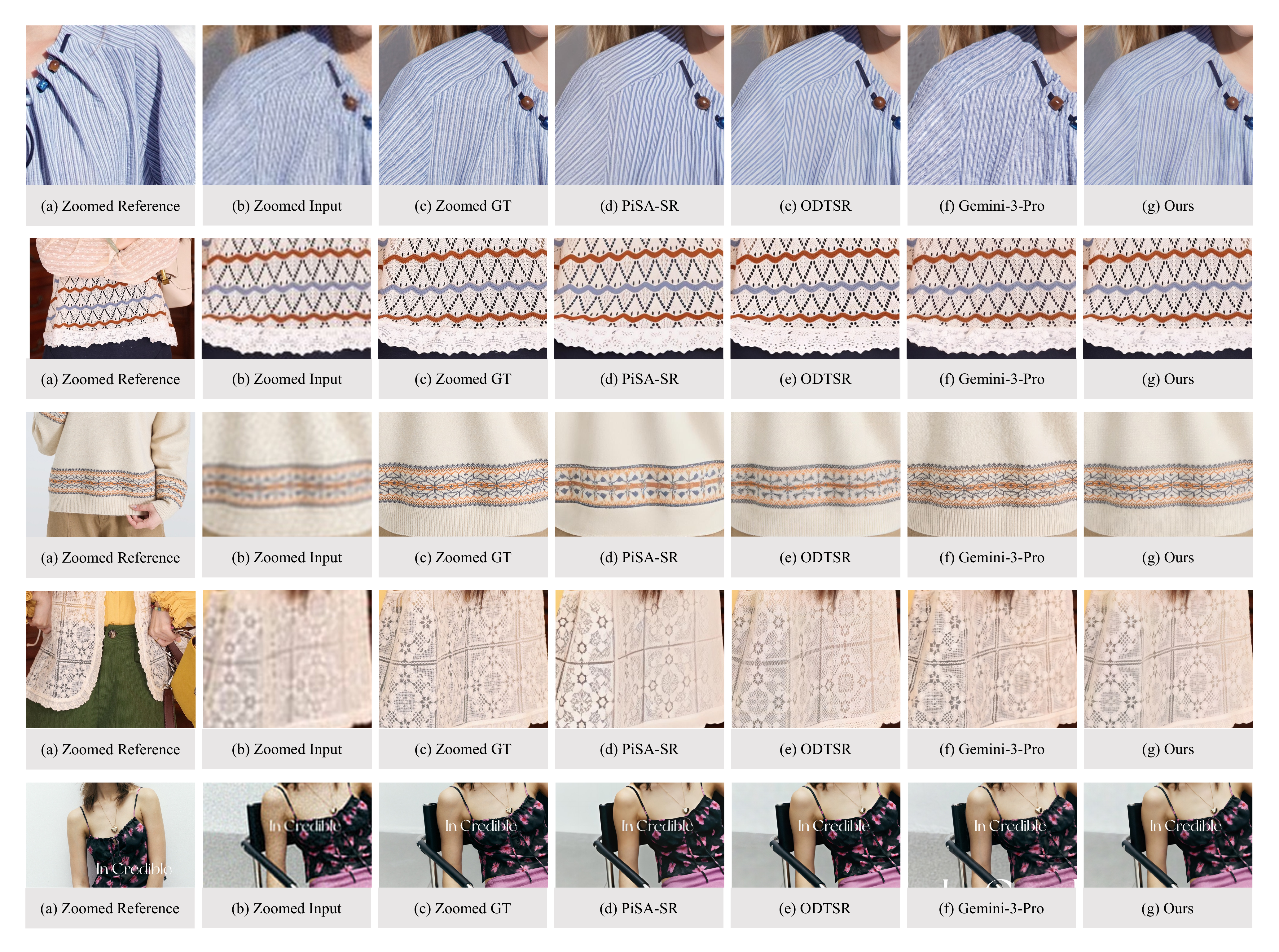}
    \caption{More qualitative comparisons on the \textbf{TexRefSR-Eval-Syn}
    dataset, where low-quality inputs are produced by the synthetic degradation model. Guided by the reference, our method restores textures that are more faithful to the ground truth, recovering fine patterns and structural details that other methods either over-smooth or hallucinate incorrectly.}
    \label{fig:vis_syn}
\end{figure*}

\begin{figure*}[!t]
    \centering
    \includegraphics[width=\linewidth]{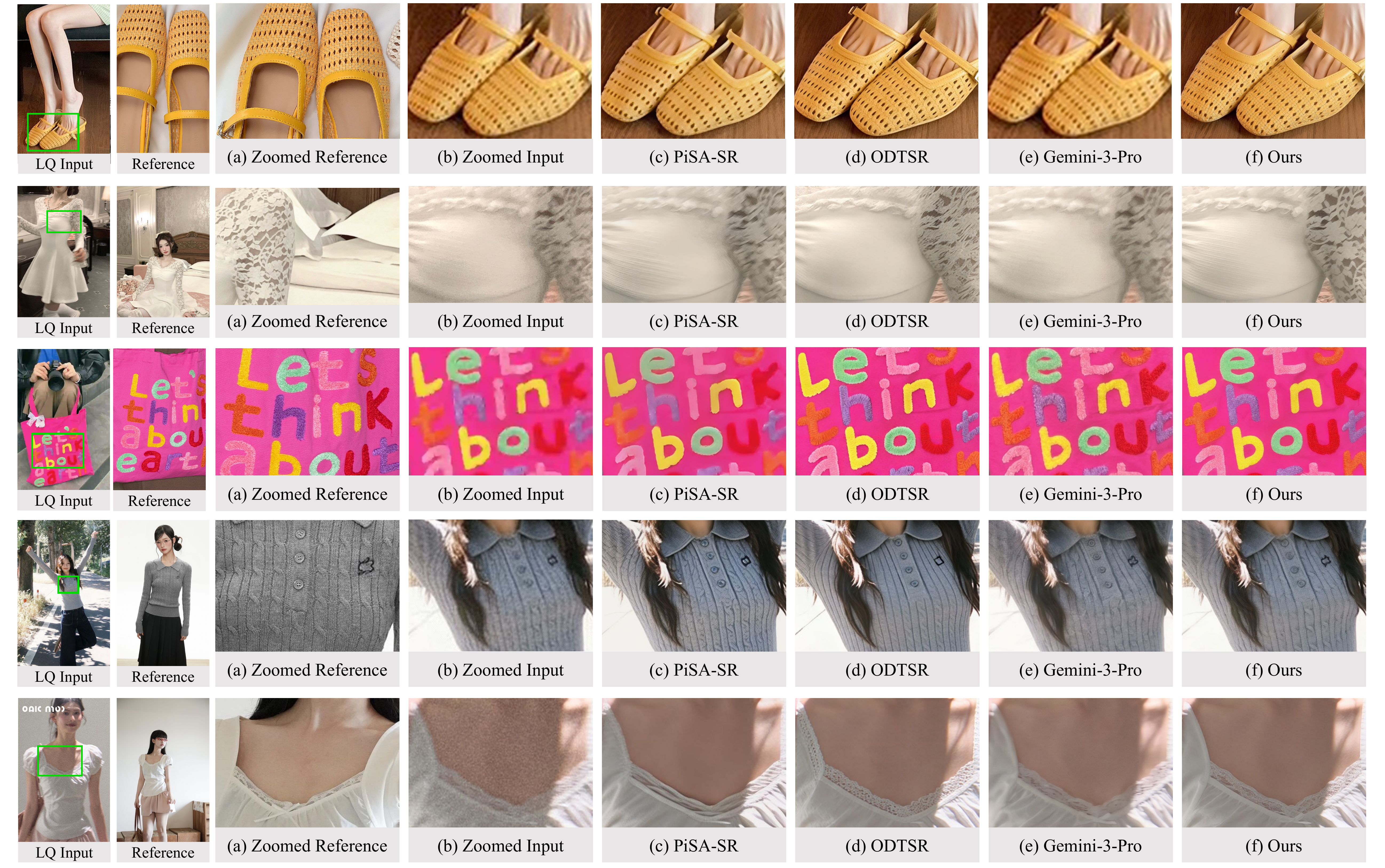}
    \caption{More qualitative comparisons on the \textbf{TexRefSR-Eval-Real} dataset, which contains real-world low-quality images without ground truth.
    Our method produces perceptually sharper results whose textures remain consistent with the reference, whereas other methods tend to yield blurry or unfaithful reconstructions.}
    \label{fig:vis_real}
\end{figure*}

\begin{figure*}[!t]
    \centering
    \includegraphics[width=\linewidth]{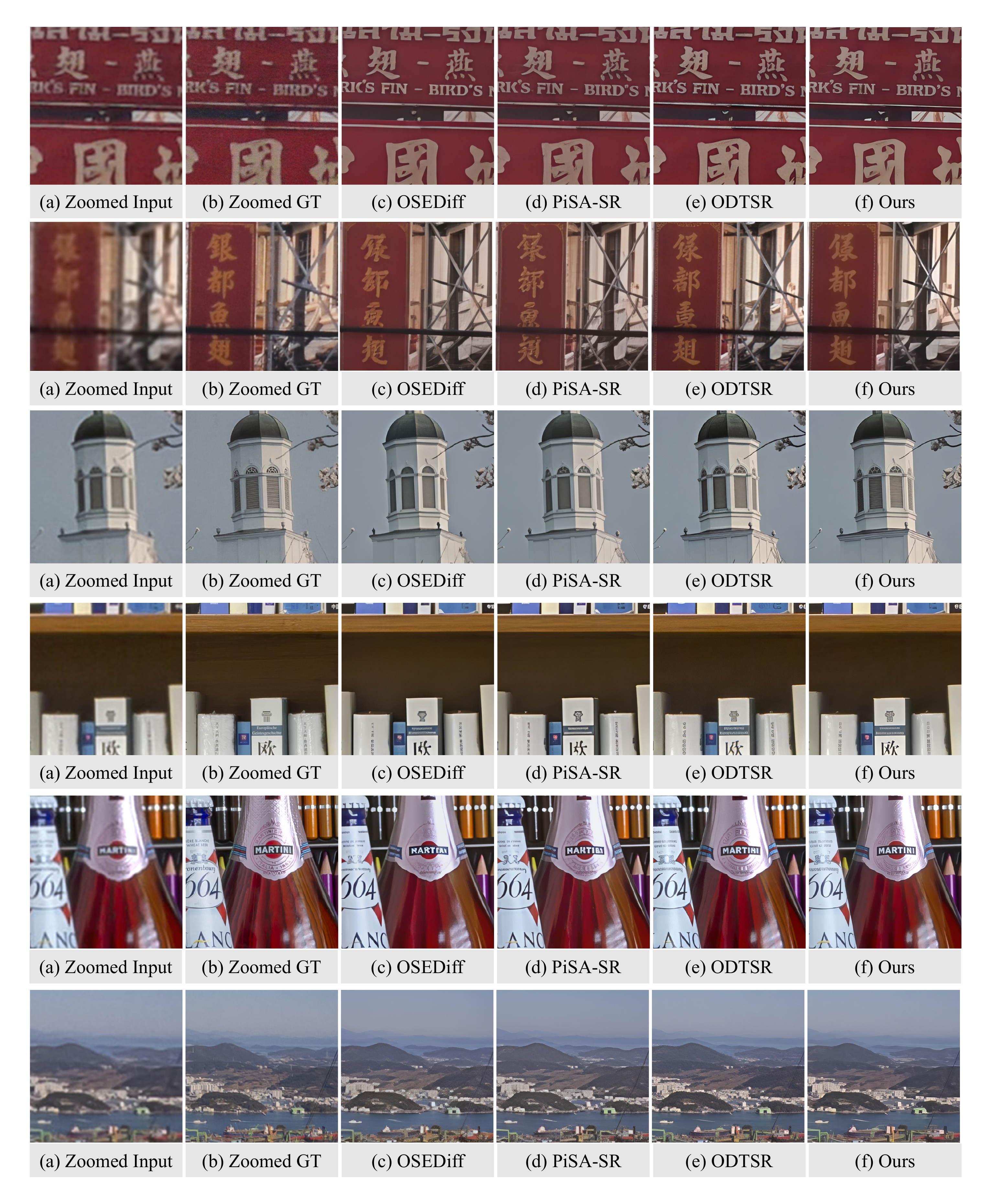}
    \caption{More qualitative comparisons on general real-world SR benchmarks (RealSR and DRealSR). Although no reference is available in this setting, our method still generalizes well to natural scenes and recovers clean, realistic details that are competitive with or superior to state-of-the-art general SR methods.}
    \label{fig:vis_general}
\end{figure*}

We provide additional qualitative comparisons to further demonstrate the effectiveness of GraftSR. Figure~\ref{fig:vis_syn} and Figure~\ref{fig:vis_real} present more results on the synthetic and real-world
splits of our TexRefSR-Eval benchmark, respectively, where reference guidance is available. Figure~\ref{fig:vis_general} further reports results on general image super-resolution benchmarks. Across all settings, our method consistently restores sharper structures and more faithful textures than competing approaches, while avoiding the over-smoothing and hallucinated details commonly observed in other methods.

\clearpage
\clearpage